\documentclass{article}
\pdfoutput=1

\usepackage[preprint, nonatbib]{neurips_data_2023}

\usepackage[utf8]{inputenc} 
\usepackage[T1]{fontenc}    
\usepackage{url}            
\usepackage{booktabs}       
\usepackage{amsfonts}       
\usepackage{nicefrac}       
\usepackage{microtype}      
\usepackage{xcolor}         
\usepackage{tabu}         
\usepackage{xspace,mfirstuc,tabulary}

\usepackage{multirow}
\usepackage{array, caption, floatrow, makecell, booktabs}
\usepackage{wrapfig}
\usepackage{floatrow}
\usepackage{comment}
\usepackage{footnote}
\usepackage{enumitem}
\usepackage{accents}

\usepackage{graphicx}
\usepackage{subfigure}
\usepackage{bbding}
\usepackage{ntheorem}
\usepackage{enumitem}
\usepackage{oplotsymbl}
\usepackage{multirow}
\usepackage{bbm}
\usepackage{adjustbox}

\usepackage{wrapfig}
\usepackage{comment}
\usepackage{enumitem}
\usepackage{accents}

\usepackage{tabu}

\usepackage{multirow,array}
\usepackage{color, colortbl}
\definecolor{Gray}{gray}{0.93}

\usepackage{booktabs}
\usepackage{pifont}
\newcommand{\cmark}{\ding{51}}%
\newcommand{\xmark}{\ding{55}}%

\usepackage[pagebackref=true,breaklinks=true,colorlinks,bookmarks=false]{hyperref}

\usepackage{floatrow}
\newlength\savewidth\newcommand\shline{\noalign{\global\savewidth\arrayrulewidth
  \global\arrayrulewidth 1pt}\hline\noalign{\global\arrayrulewidth\savewidth}}
\newcommand\paperurl[1]{{\footnotesize{\color{blue}{\url{#1}}}}}

\definecolor{rose}{HTML}{003472}

\title{\textcolor{rose}{detrex}: Benchmarking Detection Transformers}

\author{%
 \textbf{Tianhe Ren\textsuperscript{\rm 1}\thanks{Equal contributions. List order is random.} \thanks{Project lead.},\; Shilong Liu\textsuperscript{\rm 1}$^{\mathrm{*}}$,\; Feng Li\textsuperscript{\rm 1}$^{\mathrm{*}}$,\; Hao Zhang\textsuperscript{\rm 1}$^{\mathrm{*}}$,\; Ailing Zeng\textsuperscript{\rm 1},\;}\\
  \textbf{Jie Yang\textsuperscript{\rm 1},\; Xingyu Liao\textsuperscript{\rm 2},\; Ding Jia\textsuperscript{\rm 3},\; Hongyang Li\textsuperscript{\rm 1},\; He Cao\textsuperscript{\rm 1},\; Jianan Wang\textsuperscript{\rm 1},\;}\\
  \textbf{Zhaoyang Zeng\textsuperscript{\rm 1},\; Xianbiao Qi\textsuperscript{\rm 1},\; Yuhui Yuan\textsuperscript{\rm 4},\; Jianwei Yang\textsuperscript{\rm 5},\; Lei Zhang\textsuperscript{\rm 1}\thanks{Corresponding author.}}\\[0mm]
  \textsuperscript{\rm 1}International Digital Economy Academy (IDEA)\\
  \textsuperscript{\rm 2}University of Science and Technology of China \\
  \textsuperscript{\rm 3}Peking University \quad \textsuperscript{\rm 4}Microsoft Research Asia \quad \textsuperscript{\rm 5}Microsoft Research, Redmond \\
}

\begin{document}

\maketitle

\begin{abstract}
  The DEtection TRansformer (DETR) algorithm has received considerable attention in the research community and is gradually emerging as a mainstream approach for object detection and other perception tasks. However, the current field lacks a unified and comprehensive benchmark specifically tailored for DETR-based models. 
  To address this issue, we develop a unified, highly modular, and lightweight codebase called \textbf{\textcolor{rose}{detrex}}, which supports a majority of the mainstream DETR-based instance recognition algorithms, covering various fundamental tasks, including object detection, segmentation, and pose estimation. 
  We conduct extensive experiments under \textbf{\textcolor{rose}{detrex}} and perform a comprehensive benchmark for DETR-based models. 
  Moreover, we enhance the performance of detection transformers through the refinement of training hyper-parameters, providing strong baselines for supported algorithms.
  We hope that \textbf{\textcolor{rose}{detrex}} could offer research communities a standardized and unified platform to evaluate and compare different DETR-based models while fostering a deeper understanding and driving advancements in DETR-based instance recognition. Our code is available at \href{https://github.com/IDEA-Research/detrex}{https://github.com/IDEA-Research/detrex}. The project is currently being actively developed. We encourage the community to use \textbf{\textcolor{rose}{detrex}} codebase for further development and contributions.

\end{abstract}

\section{Introduction}
  Instance recognition such as object detection and instance segmentation are fundamental problems in computer vision with wide-ranging applications. In the past decade, significant progress has been made in this field. Most classical detectors rely on convolutional architectures with meticulously handcrafted components, including pre-defined anchors and Non-maximum Suppression (NMS) post-process~\cite{ren2015faster, liu2016ssd, redmon2017yolo9000, tian2019fcos, law2018cornernet, redmon2016you}. In contrast, the recently proposed DEtection TRansformer (DETR)~\cite{carion2020end} model introduces the Transformer architecture into the realm of object detection for the first time and views object detection as a set-prediction task. This eliminates the need for NMS post-process and enables fully end-to-end training for object detection, setting up a new pipeline in this field. 
  Many follow-up works have not only introduced numerous enhancements to the DETR model~\cite{zhu2020deformable, gao2021fast, meng2021conditional, wang2021anchor, liu2022dabdetr, li2022dn, zhang2022dino, chen2022group, chen2022group2, jia2022detrs, ouyang2022nms, liu2023detection}, continuously improving the performance of DETR-based detectors, but also demonstrated the versatility of DETR-based algorithms in a variety of domains such as image segmentation~\cite{cheng2022masked, cheng2021maskformer, li2022mask}, pose estimation~\cite{yang2023explicit, Shi_2022_CVPR}, and 3D detection~\cite{li2022bevformer, zhang2023bev}. In this paper, we define these DETR series or their applications in other domains as DETR-based models.

  With the rapid development of detection transformers and other DETR-based algorithms, we have observed that existing algorithms are implemented independently. There is still a lack of a unified and comprehensive benchmark to ensure the effectiveness of newly developed DETR-based methods and fair comparisons among them. The existing frameworks like MMDetection~\cite{chen2019mmdetection} and Detectron2~\cite{wu2019detectron2} face challenges in effectively benchmarking DETR-based algorithms. Specifically, the MMDetection framework categorizes traditional CNN-based one-stage and two-stage object detection algorithms into components such as \textit{Backbone}, \textit{Neck}, \textit{Head}, and \textit{Assigner}. However, due to the differences in design between DETR-based algorithms and traditional CNN-based algorithms, supporting DETR-based algorithms under the MMDetection framework requires extensive code modifications to adapt to its original modular design. This significantly increases the cost of reproducing the algorithms. On the other hand, Detectron2 framework is a highly efficient and lightweight detection framework that supports comprehensive instance recognition tasks but lacks proper modular design for DETR variants. To address this issue and to facilitate the research community and industry in the more efficient development and evaluation of DETR-based algorithms, we have designed and developed a highly modular and lightweight codebase, termed \textbf{\textbf{\textcolor{rose}{detrex}}}. 

  Within \textbf{\textcolor{rose}{detrex}}, to better assist researchers in the development of DETR-based algorithms and enhance the development efficiency, we have implemented a unified and modular design for the DETR-based models and incorporated a lightweight training engine along with a concise and effective configuration system. This allows researchers to adjust configurations and modify model structures more flexibly during training and evaluation. Furthermore, we have integrated various DETR-based models that cover various perceptual tasks, including object detection, instance/semantic/panoptic segmentation, and pose estimation, into the framework. As a result, the \textbf{\textcolor{rose}{detrex}} framework provides a comprehensive toolbox for training, validation, and benchmarking of these models. 
  
  With a well-structured and unified codebase, we have conducted extensive experiments, supporting and reproducing over 15 mainstream DETR-based detection, segmentation, and pose estimation algorithms. Notably, compared to the original implementations of these algorithms, we have further boosted the model performance ranging from \textbf{0.2 AP} to \textbf{1.1 AP} through optimizing both model and training hyper-parameters. We then conducted fair comparisons to benchmark the DETR-based algorithms, with respect to various aspects such as the model performance, training time, inference speed, the impact of different backbones on performance, and the influence of different modules within the algorithm on the final performance. 
  We summarize our main contributions as follows:

  \textbf{A Comprehensive Toolbox for DETR-based Algorithms.} We provide a unified open-source toolbox called \textbf{\textbf{\textcolor{rose}{detrex}}} with a highly modular and extensible design for DETR-based detection, segmentation, and pose-estimation algorithms. We provide a rich set of model reproductions, including Deformable-DETR~\cite{zhu2020deformable}, Conditional-DETR~\cite{meng2021conditional}, Anchor-DETR~\cite{wang2021anchor}, DAB-DETR~\cite{liu2022dabdetr}, DN-DETR~\cite{li2022dn}, DINO~\cite{zhang2022dino}, Group-DETR~\cite{chen2022group}, H-DETR~\cite{jia2022detrs}, Mask2Former~\cite{cheng2022masked}, 
  MP-Former~\cite{zhang2023mp},
  MaskDINO~\cite{li2022mask}, EDPose~\cite{yang2023explicit}, etc. In addition, we offer comprehensive documentation and tutorials to facilitate easy modification of our codebase, which is open for ongoing development. We will continue incorporating new algorithms and results into the framework to enhance its functionality.

  \textbf{Comprehensive Detection Transformer Benchmark.} We conducted a comprehensive benchmarking analysis primarily focused on DETR-based models. Firstly, we validated the effectiveness of our implemented DETR-based detection algorithms on COCO \texttt{val2017}, including the model performance, parameters, GFLOPs, FPS, and training hours. Subsequently, we evaluated the effectiveness of the latest backbones based on the DINO~\cite{zhang2022dino} detector. In addition, we conducted an analysis of the influence of different hyper-parameters and components of the model on its final performance. Furthermore, we validated the performance of DETR-based algorithms, such as Mask2Former, MaskDINO, and ED-Pose, in the domains of segmentation and pose estimation.

  \textbf{Strong Baselines for Detection Transformers.} 
  In addition to reproducing the model results based on our codebase, we also conducted extensive experiments on the hyper-parameters of each model, resulting in a significant improvement in the reproduced results. We share several findings: Non-maximum Suppression (NMS) still shows benefits in DETR variants. The impact of hyper-parameters on model performance is significant, which requires further exploration.
  As a result, our \textbf{\textcolor{rose}{detrex}} reproductions of the DAB-DETR, Deformable-DETR, DAB-Deformable-DETR, H-DETR, and DINO models have demonstrated significant performance improvements, with respective gains of 1.1AP, 1.1AP, 1.0AP, 0.5AP, and 0.7AP on COCO \texttt{val2017}, compared to their original implementations.
  
  \vspace{-2mm}
\section{Related Work}
\subsection{Detection Transformers}
\vspace{-2mm}
Object detection is a fundamental task in computer vision and has been extensively studied in the literature. 
The mainstream object detection approaches are based on deep neural networks, especially convolutional neural networks (CNN).
CNN-based detectors \cite{ren2015faster} often require the incorporation of manually designed modules, such as Non-Maximum Suppression (NMS), to post-process the network outputs.
To address these limitations, DEtection TRansformer (DETR) \cite{carion2020end} was proposed as a pioneering end-to-end solution. 
It introduces the Transformer architecture into object detection for the first time and views object detection as a set-prediction task.

Despite its novelty, DETR encountered challenges such as slow convergence speed, inferior performance compared to mainstream solutions, and an unclear understanding of query semantics.
Subsequent research efforts have focused on addressing these issues and have made significant progress. Such improvements includes enhancing attention mechanisms \cite{zhu2020deformable}, improving query interpretability \cite{meng2021conditional, liu2022dabdetr, wang2021anchor}, introducing auxiliary tasks \cite{li2022dn, nguyen2022boxer, chen2022group, chen2022group2}, and refining matching strategies \cite{ouyang2022nms, liu2023detection, ren2023strong, zhang2023dense}. Notably, DINO \cite{zhang2022dino}, which adopted a complete DETR-like structure with fewer data and parameters, achieved first place in the COCO object detection leaderboard, demonstrating a remarkable potential of DETR-based models in terms of performance.
Furthermore, recent advancements, such as RT-DETR \cite{lv2023detrs}, have showcased the practical applicability of these models by surpassing the extensively optimized YOLO series \cite{ge2021yolox, bochkovskiy2020yolov4}. Beyond standard object detection, DETR-like models have demonstrated advantages in other domains, including image segmentation \cite{hu2021istr, hu2023segment, cheng2021maskformer, li2022mask, xie2021segformer, zhang2023mp}, 3D object detection \cite{li2022bevformer, zhang2023bev}, and multimodal detection \cite{kamath2021mdetr, liu2022dq, liu2023grounding, yang2023boosting}.  As a result, DETR-based object detection has gained increasing attention from the research community and industry practitioners. 

However, lacking a well-established codebase and benchmark for DETR-based models hinders their further development and evaluation.
To close this gap, we develop a comprehensive codebase and benchmark suite \textbf{\textcolor{rose}{detrex}}, helping researchers to easily build upon and compare different approaches for DETR-based object detection. 
Through the development of \textbf{\textcolor{rose}{detrex}}, we aim to facilitate the advancement of DETR and its variants, fostering further progress in the field of object detection.
\vspace{-2mm}
\subsection{Detection Toolboxes}
\vspace{-2mm}
Over the years, the field of object detection has witnessed significant progress with the development of various detection toolboxes, such as Detectron2 \cite{wu2019detectron2} and MMDetection \cite{chen2019mmdetection}. 
These toolboxes have played a crucial role in advancing object detection research and facilitating practical applications. They provide a comprehensive set of pre-defined models, training pipelines, and evaluation metrics, making it easier for researchers and practitioners to develop and deploy object detection systems.

However, these existing repositories primarily focus on CNN-based detectors and lack specific design considerations for Transformer-based models. The \textbf{\textcolor{rose}{detrex}} codebase, on the other hand, is primarily designed for DETR-like models, providing more concise and well-structured support for Transformer-based models. 
Additionally, not only does \textbf{\textcolor{rose}{detrex}} support the object detection task, but it also covers a broader range of task categories, including instance segmentation, pose estimation, etc. 
This versatility allows \textbf{\textcolor{rose}{detrex}} to be applied to various tasks beyond object detection, expanding its applicability and utility for researchers and developers.

\section{Highlights of \textbf{\textcolor{rose}{detrex}}}

  In this section, we mainly focus on the modular design of Transformer-based algorithms and the system design within \textbf{\textcolor{rose}{detrex}}, the overall design principle of \textbf{\textcolor{rose}{detrex}}, and a comparative analysis between \textbf{\textcolor{rose}{detrex}} and other codebases.

  \begin{figure}[h]
    \centering
    \vspace{-0.2cm}
    \includegraphics[width=0.98\linewidth]{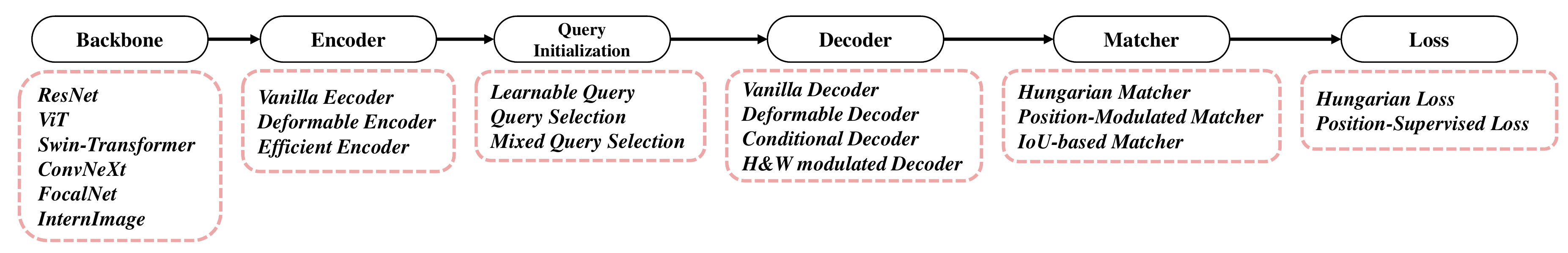}
    \vspace{-0.2cm}
    \caption{The modular design for DETR-based algorithms under detrex. }
    \vspace{-0.2cm}
    \label{fig:modular_design}
\end{figure}

  \subsection{Modular and Extensible Design for Transformer-based Models}
  Within the \textbf{\textcolor{rose}{detrex}} framework, we have devised a unified modular design for the Transformer-based detection models, which can also be seamlessly applied to segmentation and pose estimation models, resulting in a highly extensible system. We modularized the Transformer-based detection algorithms into six components, including \textit{Backbone}, \textit{Encoder}, \textit{Query Initialization}, \textit{Decoder}, \textit{Matcher} and \textit{Loss}. Our modular design is illustrated in Fig.~\ref{fig:modular_design}. For each modular, we have already incorporated multiple state-of-the-art methods as off-the-shelf options for developers. Our design also allows the smooth incorporation of user-defined modules into any system component.

  \begin{figure}[h]
    \centering
    \includegraphics[width=0.98\linewidth]{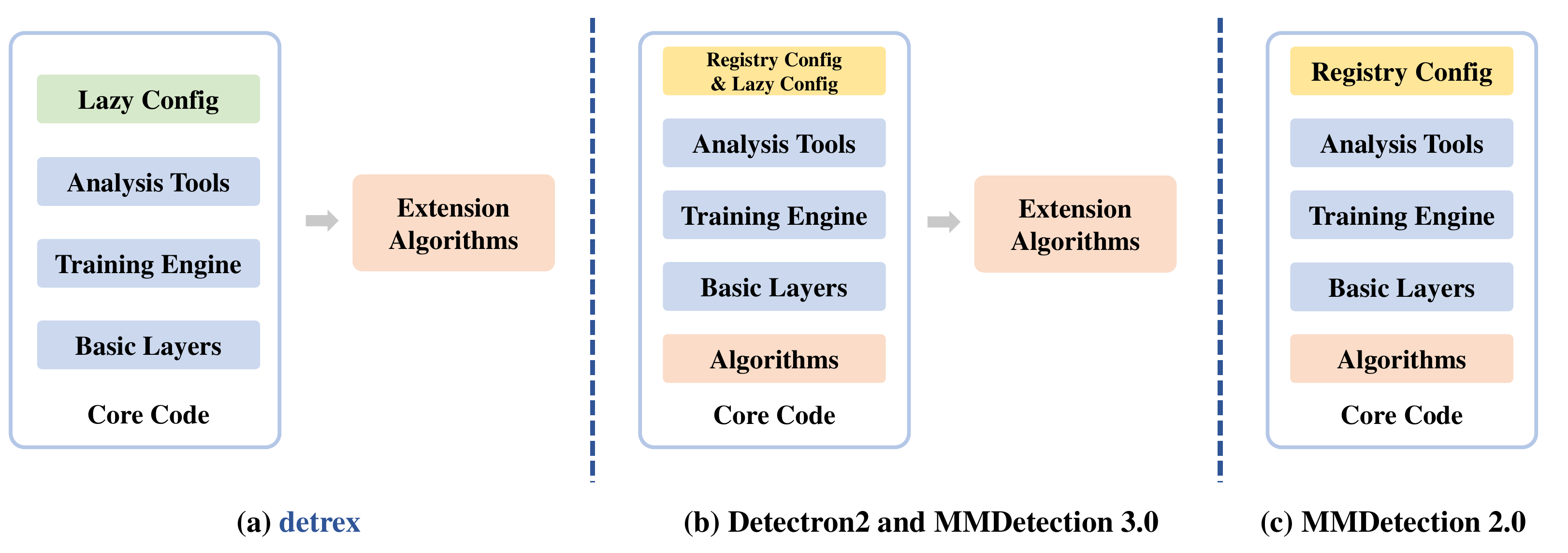}
    \vspace{-0.3cm}
    \caption{The design comparisons of \textbf{\textcolor{rose}{detrex}} with other codebases. (a) In \textbf{\textcolor{rose}{detrex}}, we opted for a lightweight and flexible approach by exclusively employing LazyConfig configuration system and maintaining all supported algorithms as separate extension projects. (b) MMDetection 3.0 and Detectron2 support the users to implement algorithms outside the core code. Detectron2 supports both the Registry Config and a more flexible LazyConfig. (c) MMDetection 2.0 implemented all core algorithms within the core code and utilized a Configuration System based on the registration mechanism. 
    }
    \label{fig:codebase_comparison}
\end{figure}
  \vspace{-0.4cm}
  \subsection{Benchmarking Toolbox Design Principle}
  The overall architecture of \textbf{\textcolor{rose}{detrex}} is illustrated in Fig.\ref{fig:codebase_comparison} (a), and our design principles for \textbf{\textcolor{rose}{detrex}} can be summarized as follows:

  \textbf{Highly flexible configuration system}. Under the \textbf{\textcolor{rose}{detrex}} framework, we fully adopted the \textbf{LazyConfig} configuration system, which was developed in the latest version of Detectron2. In contrast to the traditional approach of utilizing registry mechanisms and string-based configuration in MMDetection and the old version of Detectron2. LazyConfig employs a non-intrusive design and enables users to define its configuration, even complex data types or objects, entirely using Python syntax. Moreover, The entirety of the configuration content is notably \textbf{concise}, \textbf{clean}, and \textbf{lightweight}. This affords users great convenience in customizing the configurations to their specific needs.

  \textbf{Advanced training techniques}. \textbf{\textcolor{rose}{detrex}} supports a comprehensive set of training techniques, including {Mixed Precision Training}, {Activation Checkpointing}, {Exponential Moving Average}, Distributed Data-Parallel Training, etc. These techniques can help users effectively train their models under resource-constrained environments.

  \textbf{Effective project management practices}. \textbf{\textcolor{rose}{detrex}} utilizes a project-based management approach, where each algorithm is treated as an independent project, maintained separately from the core code. This decoupling of projects and core code ensures that algorithms are self-contained and unaffected by changes in the core code. Moreover, by combining with the modular design, the project-based approach can effectively guarantee the framework's \textbf{maintainability}.

  \textbf{Rich and practical analysis tools}. \textbf{\textcolor{rose}{detrex}} supports a rich set of practical tools for users to analyze the existing algorithms or their models, including counting FLOPs, testing inference speed, visualizing predictions during training times, or visualizing the inference results. 

  \textbf{Support for a range of perceptual tasks and datasets}. \textbf{\textcolor{rose}{detrex}} not only supports detection tasks but also enables semantic/instance/panoptic segmentation and pose estimation tasks. It provides compatibility with popular datasets, including COCO~\cite{lin2014microsoft}, ADE20k~\cite{zhou2019semantic}, Cityscapes~\cite{cordts2016cityscapes}, Mapillary Vistas~\cite{neuhold2017mapillary}, as well as large-scale datasets like Object-365~\cite{shao2019objects365}, which allows users to verify model performance across different datasets and also facilitates scaling up model training to further enhance its performance.

  \subsection{Comparison between \textbf{\textcolor{rose}{detrex}} and Previous Frameworks}
  \begin{table*}[ht]
\caption{{Comparison between \textbf{\textcolor{rose}{detrex}} and other related frameworks.}}
\vspace{-0.3cm}
\centering\setlength{\tabcolsep}{7pt}
\renewcommand{\arraystretch}{1.2}
\footnotesize
\vspace{-0mm}
\setlength{\tabcolsep}{15pt}
\renewcommand{\arraystretch}{1.2}
\resizebox{1\columnwidth}{!}{%
\begin{tabular}{ccccccc}
\shline
Benchmark & \# Supported DETRs  & \# Tasks & Lightweight  & Maintainability & Focus on DETR \\
\shline
MMDetection 2.0~\cite{chen2019mmdetection} & 4  & 3 & \xmark  & \xmark & \xmark \\
MMDetection 3.0~\cite{chen2019mmdetection} & 7  & 4 & \xmark  & \cmark & \xmark \\
Detectron2~\cite{wu2019detectron2}  & 3  & \textbf{5} & \cmark  & \cmark & \xmark \\
detrex (ours)  & \textbf{19} & \textbf{5}  & \cmark & \cmark  & \cmark \\
\shline
\end{tabular}
\vspace{-0.4cm}
\label{tab:benchmark_comparison}}
\end{table*}

  We compare \textbf{\textcolor{rose}{detrex}} and previous codebases in Table~\ref{tab:benchmark_comparison} and Fig.~\ref{fig:codebase_comparison}, where \textbf{\textcolor{rose}{detrex}} offers distinct advantages for the development of Transformer-based algorithms: \textbf{\textcolor{rose}{detrex}} provides unified modular design specific for Transformer-based models that allow researchers to incorporate their algorithms into any part of the framework easily. And in contrast to the design of MMDetection 2.0, MMDetection 3.0, and Detectron2, where certain implemented algorithms are integrated into the core code, \textbf{\textcolor{rose}{detrex}} follows a more \textbf{lightweight and maintainable} approach by fully decoupling the core code from the algorithm projects. \textbf{\textcolor{rose}{detrex}} also provides extensive support not only for a wide range of Transformer-based algorithms but also for various perception tasks, including object detection, semantic/instance/panoptic segmentation, and pose estimation. Furthermore, we have conducted extensive experiments within the \textbf{\textcolor{rose}{detrex}} framework, which yields improved model reproducibility. This enables researchers to conduct more effective comparisons between Transformer-based algorithms.

\begin{table*}[ht]
\centering\setlength{\tabcolsep}{5pt}
\renewcommand{\arraystretch}{1.5}
\footnotesize
\caption{Benchmarking the performance of DETR variants with a ResNet-50 backbone on COCO \texttt{val2017}. The best and second-best results are highlighted in bold and underlined, respectively.}
\vspace{-0.2cm}
{
\resizebox{1.0\textwidth}{!}{%
\begin{tabular}{l|c|c|ccccc|c|c|c|c|c}
\shline
\textbf{Model}  & \textbf{\#ep} & \textbf{AP} & \textbf{AP$_{50}$} & \textbf{AP$_{75}$} & \textbf{AP$_{S}$} & \textbf{AP$_{M}$} & \textbf{AP$_{L}$} & \textbf{\#params} & \textbf{GFLOPs} & \textbf{FPS} & \textbf{Memory} & \textbf{GPU-h} \\
\shline
Deformable-DETR-Two-Stage~\cite{zhu2020deformable}  & 50  & $48.2$ & 67.0 & 52.2 & 30.7 & 51.4 & 63.0 & 41.2M & 175.6 ± 19.1 & 26.3 & 11.0GB & 208h \\
Anchor-DETR~\cite{wang2021anchor}  & 50  & $41.9$ & 62.9 & 44.6 & 22.0 & 46.0 & 59.7 & 37.0M & 92.7 ± 9.2 & 27.8 & 44.7GB & 168h\\
Conditional-DETR~\cite{meng2021conditional}  & 50  & $41.6$ & 63.0 & 43.9 & 21.4 & 45.2 & 59.8 & 43.4M & 89.1 ± 9.7 & 37.8 & 6.4GB & 164h \\
DAB-DETR~\cite{liu2022dabdetr}  & 50   & $43.3$ & 63.9 & 45.9 & 23.4 & 47.1 & 62.1 & 43.7M & 90.4 ± 9.7 & 32.9 & 5.0GB & 214h \\
DN-DETR~\cite{li2022dn}  & 50   & $44.7$ & 65.3 & 47.5 & 23.7 & 48.7 & 64.1 & 43.7M & 90.5 ± 9.7 & 32.2 & 5.1GB & 240h \\
DAB-Deformable-DETR~\cite{liu2022dabdetr}  & 50   & \underline{49.0} & \underline{67.4} & \underline{53.4} & \underline{31.5} & \underline{52.1} & \underline{64.4} & 47.4M & 231.3 ± 25.1 & 23.4 & 10.5GB  & 230h \\
DAB-Deformable-DETR-Two-Stage~\cite{liu2022dabdetr}  & 50   & \textbf{49.7} & \textbf{68.0} & \textbf{54.3} & \textbf{31.9} & \textbf{53.2} & \textbf{64.7} & 47.5M & 235.4 ± 255 & 22.1 & 10.5GB & 220h\\
\hline
DINO-4scale~\cite{zhang2022dino}  & 12   & \underline{49.7} & 67.0 & \underline{54.4} & 31.4 & \underline{52.9} & 63.6 & 47.7M & 244.5 ± 25.5 & 24.6 & 10.9GB & 67h \\
$\mathcal{H}$-DETR~\cite{jia2022detrs}  & 12   & $49.1$ & 66.9  & 53.7 & 32.2 & 52.3 & 63.8 & 47.9M & 268.1 ± 24.7 & 22.4 & 12.0GB & 80h \\
DETA-5scale~\cite{ouyang2022nms} & 12   & \textbf{50.2} & \textbf{67.4} & \textbf{55.2} & \underline{32.3} & \textbf{54.2} & \textbf{65.0} & 48.4M & 247.1 ± 25.9 & 15.3 & 10.8GB & 53h \\
\shline
\end{tabular}}
\label{tab:compare_with_r50_backbone}
}
\end{table*}
  \begin{table*}[ht]
\vspace{-0.2cm}
\centering\setlength{\tabcolsep}{3pt}
\renewcommand{\arraystretch}{1.5}
\footnotesize
\vspace{-0mm}
{
\resizebox{1.0\textwidth}{!}{%
\begin{tabular}{l|l|l|cccccc|c|c|c|c|c}
\shline
\textbf{Backbone} & \textbf{Pretrained}  & \textbf{\#ep}  & \textbf{AP} & \textbf{AP$_{50}$} & \textbf{AP$_{75}$} & \textbf{AP$_{S}$} & \textbf{AP$_{M}$} & \textbf{AP$_{L}$} & \textbf{\#params} & \textbf{GFLOPs} & \textbf{FPS} & \textbf{Memory} & \textbf{GPU-h} \\
\shline
\multicolumn{14}{l}{\textit{\cellcolor{Gray}\textbf{ResNet Backbone}}} \\
R50 & IN-1K  & 12   & $49.7$ & 67.0 & \underline{54.4} & 31.4 & 52.9 & 63.6 & 47.7M & 244.5 ± 25.5 & 24.6 & 10.9GB & 67h \\
R50 & IN-1K  & 24   & \textbf{50.6} & \textbf{68.4} & \textbf{55.4} & \textbf{34.9} & \textbf{53.8} & \textbf{65.0} & 48.0M & 244.5 ± 25.5 & 24.6 & 10.9GB & 132h \\
R101 & IN-1K  & 12   & \underline{50.0} & \underline{67.7} & \underline{54.4} & \underline{32.4} & \underline{53.4} & \underline{64.2} & 66.6M & 310.8 ± 32.9 & 19.3 & 12.8GB & 73h \\
\shline
\multicolumn{14}{l}{\textit{\cellcolor{Gray}\textbf{Swin-Transformer Backbone}}} \\
Swin-Tiny & IN-1K & 12   & $51.3$ & 69.0 & 56.0 & 34.5 & 54.4 & 66.0 & 48.2M & 252.3 ± 25.9 & 20.5 & 14.4GB & 107h\\
Swin-Tiny & IN-22K & 12   & $52.5$ & 70.6 & 57.3 & 35.4 & 55.7 & 68.2 & 48.2M & 252.3 ± 25.9 & 20.5 & 14.4GB & 107h\\
Swin-Small & IN-1K & 12   & $53.0$ & 71.2 & 57.6 & 36.0 & 56.7 & 68.3 & 69.5M & 332.1 ± 34.0 & 17.3 & 19.9GB & 128h\\
Swin-Small & IN-22K & 12   & $54.5$ & 72.7 & 59.5 & 38.1 & 57.4 & 70.6 & 69.5M & 332.1 ± 34.0 & 17.3 & 19.9GB & 128h\\
Swin-Base & IN-1K & 12   & $53.4$ & 71.8 & 58.3 & 35.9 & 56.7 & 69.6 & 108.4M & 484.0 ± 49.6 & 15.6 & 27.9GB & 158h\\
Swin-Base & IN-22K & 12   & $55.8$ & 74.3 & 60.7 & \underline{39.0} & 59.2 & 72.8 & 108.4M & 484.0 ± 49.6 & 15.6 & 27.9GB & 158h\\
Swin-Large & IN-22K  & 12   & \underline{56.9} & \underline{75.7} & \underline{62.2} & 38.8 & \underline{61.0} & \underline{73.6} & 217.8M & 813.9 ± 83.8 & 12.9 & 38.0GB & 178h \\
Swin-Large & IN-22K  & 36   & \textbf{58.1} & \textbf{77.1} & \textbf{63.5} & \textbf{41.2} & \textbf{62.1} & \textbf{73.9} & 217.8M & 813.9 ± 83.8 & 12.9 & 38.0GB & 320h \\
\shline
\multicolumn{14}{l}{\textit{\cellcolor{Gray}\textbf{FocalNet Backbone}}} \\
Focal-Tiny-3level & IN-1K & 12  & 53.0 & 71.1 & 57.9 & 36.2 & 56.4 & 67.8 & 48.6M & 248.6 ± 26.0 & 20.4 & 16.8GB & 77h \\
Focal-Small-3level & IN-1K  & 12  & 54.1 & 72.4 & 59.3 & 37.3 & 57.7 & 69.4 & 70.2M & 324.4 ± 34.4 & 16.9 & 22.5GB & 100h \\
Focal-Base-3level & IN-1K & 12  & 54.4 & 72.9 & 59.3 & 37.7 & 57.7 & 69.6 & 109.1M & 444.2 ± 47.2 & 15.4 & 26.4GB & 108h \\
Focal-Large-3level & IN-22K  & 12   & $57.5$ & 76.0 & 62.8 & \textbf{41.8} & 61.3 & 73.2 & 228.3M & 807.4 ± 87.8 & 11.4 & 36.7GB & 132h \\
Focal-Large-4level & IN-22K  & 12   & \underline{58.0} & \underline{76.6} & \underline{63.4} & \underline{41.3} & \textbf{62.0} & \textbf{74.3} & 228.5M & 808.2 ± 87.9 & 10.5 & 44.0GB & 137h \\
Focal-Large-3level & IN-22K  & 36   & \textbf{58.3} & \textbf{77.4} & \textbf{63.6} & \underline{41.3} & \underline{61.9} & \underline{73.8} & 228.3M & 807.4 ± 87.8 & 11.4 & 36.7GB & 200h \\
\shline
\multicolumn{14}{l}{\textit{\cellcolor{Gray}\textbf{Vision-Transformer Backbone}}} \\
ViT-Base & IN-1K, MAE  & 12   & $50.2$ & 68.1 & 54.6 & 31.9 & 54.0 & 65.4 & 108.2M & 578.9 ± 69.3 & 13.5 & 29.1GB & 96h \\
ViT-Base & IN-1K, MAE  & 50   & \underline{55.0} & \underline{73.3} & \underline{60.3} & \underline{37.3} & \underline{58.9} & \underline{70.2} & 108.2M & 578.9 ± 69.3 & 13.5 & 29.1GB & 202h\\
ViT-Large & IN-1K, MAE & 12  & $52.9$ & 71.1 & 57.5 & 35.9 & 56.5 & 67.4 & 326.8M & 1433.6 ± 166.5 & 8.8 & 55.9GB & 140h \\
ViT-Large & IN-1K, MAE  & 50   & \textbf{57.5} & \textbf{75.9} & \textbf{63.1} & \textbf{40.9} & \textbf{60.8} & \textbf{72.4} & 326.8M & 1433.6 ± 166.5 & 8.8 & 55.9GB & 578h\\
\shline
\multicolumn{14}{l}{\textit{\cellcolor{Gray}\textbf{ConvNeXt Backbone}}} \\
ConvNeXt-Tiny & IN-1K  & 12  & 51.4 & 70.0 & 56.0 & 32.8 & 55.0 & 66.1 & 48.5M  & 245.6 ± 25.4 & 21.5 & 16.3GB & 79h \\
ConvNeXt-Tiny & IN-22K  & 12  & 52.4 & 70.9 & 57.2 & 34.2 & 55.5 & 67.5 & 48.5M  & 245.6 ± 25.4 & 21.5 & 16.3GB & 79h \\
ConvNeXt-Small & IN-1K  & 12  & 52.0 & 70.9 & 56.6 & 34.4 & 55.3 & 67.4 & 70.1M & 320.0 ± 33.5 & 19.9 & 19.6GB & 83h \\
ConvNeXt-Small & IN-22K  & 12  & 54.2 & 72.8 & 59.0 & 37.3 & 57.4 & 70.0 & 70.1M & 320.0 ± 33.5 & 19.9 & 19.6GB & 83h \\
ConvNeXt-Base & IN-1K  & 12  & 52.6 & 71.7 & 57.3 & 35.9 & 56.4 & 67.8 & 108.9M & 437.9 ± 46.4 & 17.1 & 25.4GB & 89h \\
ConvNeXt-Base & IN-22K  & 12  & \underline{55.1} & \underline{74.2} & \underline{60.0} & \underline{38.1} & \underline{58.9} & \textbf{70.7} & 108.9M & 437.9 ± 46.4 & 17.1 & 25.4GB & 89h \\
ConvNeXt-Large & IN-1K  & 12  & $53.4$ & 72.3 & 58.3 & 35.9 & 57.3 & 68.6 & 219.0M & 773.2 ± 83.1 & 14.5 & 33.4GB & 104h \\
ConvNeXt-Large & IN-22K  & 12  & \textbf{55.5} & \textbf{74.6} & \textbf{60.4} & \textbf{38.8} & \textbf{59.3} & \underline{70.6} & 219.0M & 773.2 ± 83.1 & 14.5 & 33.4GB & 104h \\
\shline
\multicolumn{14}{l}{\textit{\cellcolor{Gray}\textbf{InternImage Backbone}}} \\
InternImage-Tiny & IN-1K & 12  & 52.3 & 70.2 & 57.1 & 35.3 & 55.4 & 67.1 & 48.7M & 253.7 ± 26.5 & 17.9 & 19.8GB & 98h \\
InternImage-Small & IN-1K  & 12  & 53.6 & 71.8 & 58.4 & 37.3 & 56.8 & 68.1 & 68.8M & 314.3 ± 33.3 & 17.3 & 24.9GB & 110h \\
InternImage-Base & IN-1K & 12  & \underline{54.7} & \underline{72.9} & \underline{59.8} & \underline{38.6} & \underline{58.5} & \underline{69.9} & 115.9M & 454.2 ± 48.7 & 15.7 & 30.5GB & 119h \\
InternImage-Large & IN-22K  & 12  & \textbf{57.0} & \textbf{75.8} & \textbf{62.1} & \textbf{40.8} & \textbf{60.5} & \textbf{72.5} & 241.0M & 818.0 ± 89.0 & 12.4 & 47.8GB & 144h\\
\shline
\end{tabular}
}
\caption{Comparisons of the effectiveness of various backbones based on DINO-4scale detector.}
\vspace{-0.2cm}
\label{tab:benchmark_backbone_on_dino}
}
\end{table*}
\section{Benchmarking Transformer-Based Models}
  In this section, we first conducted a comprehensive benchmark of DETR-based models using the standard ImageNet-1K pre-trained ResNet-50~\cite{he2015deep} backbone from four aspects, including the model performance, training cost, training GPU memory usage, and FPS. Then, we benchmark the newly proposed CNN-based model (ConvNeXt~\cite{liu2022convnet}, FocalNet~\cite{yang2022focal}, InternImage~\cite{wang2022internimage}) and vision transformer models (ViT~\cite{li2022exploring}, Swin-Transformer~\cite{liu2021Swin}) based on DINO~\cite{zhang2022dino} detector. Finally, we verify the effectiveness of the segmentation and pose recognition algorithms supported by \textbf{\textcolor{rose}{detrex}}.

  \vspace{-0.2cm}
  \paragraph{Dataset and implementation details.} All the experiments were conducted on the COCO \texttt{train2017} split and evaluated on COCO \texttt{val2017} split, containing 118k training images and 5k validation images repectively. If not specified, we adopt the following training settings. For the standard 12-epoch (1x) training, we train a model with AdamW~\cite{loshchilov2019decoupled} optimizer for 90k iteration with the initial learning rate being 1e-4 and a mini-batch of 16 images. The learning rate is reduced by a factor of 10 at iteration 80K. And for standard 24-epochs (2x) and 36-epochs (3x) training, the total iterations were set to 180k and 270k, respectively. And the learning rate is reduced by a factor of 10 at 150k and 225k, respectively. The architectures of all models were aligned with their original implementations, including the number of layers in the Transformer encoder and decoder, as well as the number of object queries within the model. All the models were trained using the same data augmentation strategy as employed by DETR~\cite{carion2020end}, which involves resizing the input image such that the shortest side is at least 480 and at most 800 pixels while the longest is at most 1333. Additionally, the input image is randomly cropped to a random rectangular patch, then resized again to 800-1333 with a probability of 0.5. The inference speed is tested on a single NVIDIA-A100 GPU.

  \vspace{-0.4cm}
  \paragraph{Benchmarking Transformer-based detection models.} We benchmark various DETR-based  detection algorithms on COCO \texttt{val2017}. And we evaluate all the methods with the standard ImageNet-1K pre-trained ResNet-50~\cite{he2015deep} backbone and benchmark their training efficiency, inference speed, model performance, GPU memory usage, and training time, as shown in Table~\ref{tab:compare_with_r50_backbone}.

  Firstly, regarding training efficiency, DINO and DETA display the fastest convergence speed, making them favorable candidates when quick model training is desired. Secondly, considering inference efficiency, Conditional-DETR demonstrates the fastest inference speed coupled with the least GPU memory usage, making it highly advantageous in real-time applications. As for performance, DINO and DETA excel by achieving $49.7$ AP and $50.2$ AP, respectively, under the standard 1x settings, with DINO maintaining an impressive near real-time inference speed. 

  \vspace{-0.2cm}
  \paragraph{Benchmarking backbones.} To facilitate a fair comparison of the effectiveness of different newly proposed backbones on the transformer-based detection algorithms, we employed DINO as the default model. The DINO detector currently represents the state-of-the-art performance on the COCO leaderboard, achieving remarkable AP scores of 63.1, 64.6, and 65.0 when utilizing Swin-Transformer-Large~\cite{liu2021Swin}, FocalNet-Huge~\cite{yang2022focal, ren2023strong}, and InternImage-Huge~\cite{wang2022internimage} as the respective backbones. All the results are presented in Table~\ref{tab:benchmark_backbone_on_dino}. The backbone we have benchmarked include ResNet~\cite{he2015deep}, FocalNet~\cite{yang2022focal}, Swin-Transformer~\cite{liu2021Swin}, ViT~\cite{li2022exploring}, ConvNeXt~\cite{liu2022convnet}, and InternImage~\cite{wang2022internimage}.

  Table~\ref{tab:benchmark_backbone_on_dino} reports the performance of the DINO-4scale model trained with different backbone architectures. Across the experimental results, we observe that backbones pretrained with larger datasets contribute to superior model performance. For instance, utilizing the ImageNet-22K pretrained Swin-Transformer-Tiny backbone yields a 1.2 AP improvement over the model using the ImageNet-1K pretrained backbone. In addition, using larger backbones can effectively improve the final performance. And using ViT backbone requires a longer convergence time compared to other backbones. It takes approximately 50 epochs to achieve satisfactory performance. Among the recently proposed backbones, FocalNet achieves the best performance, achieving 58.0 AP under the standard 1x settings.

  \vspace{-0.2cm}
  \paragraph{Benchmarking on segmentation and pose estimation tasks.}
  Moving beyond the object detection task, we broaden our benchmarking scope to encompass additional tasks. Specifically, we have incorporated methods such as Mask2Former~\cite{cheng2022masked}, MP-Former~\cite{zhang2023mp}, and MaskDINO~\cite{li2022mask} for instance segmentation, and ED-Pose~\cite{yang2023explicit} for pose estimation. In practice, we currently only support inference and evaluation for these methods. By leveraging the checkpoints provided by the original authors, users can reliably reproduce the official results within \textbf{\textcolor{rose}{detrex}}. Table~\ref{tab:segmentation_transformers} presents the evaluation results of these methods on COCO instance segmentation and pose estimation tasks, confirming their exact alignment with the results reported in the original papers.
    \begin{table*}[ht]
\vspace{-0.2cm}
\centering\setlength{\tabcolsep}{8pt}
\renewcommand{\arraystretch}{1.5}
\footnotesize
\vspace{-0mm}
{
\resizebox{1\textwidth}{!}{%
\begin{tabular}{l|c|c|c|cccc|ccc}
\shline
\textbf{Model}  &\textbf{backbone}&\textbf{\# query}& \textbf{\#ep}  & \textbf{AP} & \textbf{AP$_{S}$} & \textbf{AP$_{M}$} & \textbf{AP$_{L}$} & \textbf{\#params}.& \textbf{GFLOPs} & \textbf{FPS}\\
\shline
\multicolumn{11}{l}{\textit{\cellcolor{Gray}\textbf{Instance Segmentation}}} \\
Mask2Former & R50 &100 queries &50&43.7 &23.4& 47.2 &64.8 &44M &226G &9.7  \\
MP-Former &R50 &100 queries &50 &44.8 &24.7 &48.1& 65.5 &44M &226G &9.7 \\
MaskDINO & R50& 100 queries&50& 45.4& 25.2& 48.3& 65.8 & 52M &280G& 15.2\\
MaskDINO$^*$ & R50 &300 queries&50& 46.3 & 26.1& 49.3& 66.1 & 52M&286G & 14.2\\ \shline
\multicolumn{11}{l}{\textit{\cellcolor{Gray}\textbf{Pose Estimation}}} \\
ED-Pose &R50 &100 queries &50 &71.7& - &66.2 &79.7 &48M  &  258G & 18.0\\ 
\shline
\end{tabular}}
\vspace{-0.2cm}
\caption{The performance of various detection transformer-based methods on COCO instance segmentation and pose estimation tasks, evaluated by \textbf{\textcolor{rose}{detrex}}. $^*$ denotes the model uses mask-enhanced box initialization~\cite{li2022mask}. Note that the pose estimation task only considers $\rm{AP}_M$ (medium object) and $\rm{AP}_L$ (large object) as small objects in the COCO dataset are not annotated with keypoints. }
\label{tab:segmentation_transformers}
}
\end{table*}

  \begin{table*}[ht]
\centering\setlength{\tabcolsep}{7pt}
\renewcommand{\arraystretch}{1.3}
\footnotesize
\caption{Ablation study on DETR variants with NMS post-processing. We set the default NMS threshold to 0.8.}
\vspace{-0.25cm}
{
\resizebox{1.0\textwidth}{!}{%
\begin{tabular}{l|c|c|c|ccccc}
\shline
\textbf{Model}  & \textbf{\#ep} & \textbf{with NMS} & \textbf{AP} & \textbf{AP$_{50}$} & \textbf{AP$_{75}$} & \textbf{AP$_{S}$} & \textbf{AP$_{M}$} & \textbf{AP$_{L}$} \\
\shline
DETR  & 500  &  & $42.0$ & 62.3 & 44.3 & 20.7 & 45.9 & 61.1  \\
 \rowcolor{Gray}
DETR  & 500  & \cmark & $42.1$ \textcolor{red}{(+0.1)} & 62.5 \textcolor{red}{(+0.2)} & 44.4 \textcolor{red}{(+0.1)} & 20.7 & 45.9 & 61.3 \textcolor{red}{(+0.2)} \\
\shline
Deformable-DETR  & 50  &  & $48.2$ & 67.0 & 52.2 & 30.7 & 51.4 & 63.0  \\
 \rowcolor{Gray}
Deformable-DETR  & 50  & \cmark & $48.3$ \textcolor{red}{(+0.1)} & 67.3 \textcolor{red}{(+0.3)} & 52.3 \textcolor{red}{(+0.1)} & 30.7 & 51.4 & 63.1 \textcolor{red}{(+0.1)} \\
\shline
DAB-DETR  & 50  &  & $43.3$ & 63.9 & 45.9 & 23.4 & 47.1 & 62.1  \\
 \rowcolor{Gray}
DAB-DETR  & 50  & \cmark & $43.3$ & 64.0 \textcolor{red}{(+0.1)} & 45.9 & 23.4 & 47.1 & 62.2 \textcolor{red}{(+0.1)} \\
\shline
DAB-Deformable-DETR  & 50  &  & $49.0$ & 67.4 & 53.4 & 31.5 & 52.1 & 64.4  \\
 \rowcolor{Gray}
DAB-Deformable-DETR  & 50  & \cmark & $49.1$ \textcolor{red}{(+0.1)} & 67.6 \textcolor{red}{(+0.2)} & 53.5 \textcolor{red}{(+0.1)} & 31.5 & 52.1 & 64.6 \textcolor{red}{(+0.2)} \\
\shline
$\mathcal{H}$-DETR  & 12  &  & $49.1$ & 66.9 & 53.7 & 32.2 & 52.3 & 63.8  \\
 \rowcolor{Gray}
$\mathcal{H}$-DETR  & 12  & \cmark & $49.2$ \textcolor{red}{(+0.1)} & 67.3 \textcolor{red}{(+0.4)} & 53.9 \textcolor{red}{(+0.2)} & 32.2 & 52.4 \textcolor{red}{(+0.1)} & 64.1 \textcolor{red}{(+0.3)} \\
\shline
DINO-4scale  & 12  &  & $49.7$ & 67.0 & 54.4 & 31.4 & 52.9 & 63.6  \\
 \rowcolor{Gray}
DINO-4scale  & 12  & \cmark & $49.9$ \textcolor{red}{(+0.2)} & 67.5 \textcolor{red}{(+0.5)} & 54.7 \textcolor{red}{(+0.3)} & 31.5 \textcolor{red}{(+0.1)} & 53.1 \textcolor{red}{(+0.2)} & 63.9 \textcolor{red}{(+0.2)} \\
\shline
\end{tabular}}
\label{tab:ablation_nms}
}
\end{table*}
  \begin{table*}[ht]
\centering\setlength{\tabcolsep}{15pt}
\renewcommand{\arraystretch}{1.2}
\footnotesize
\vspace{-0.25cm}
\caption{Ablation study on DINO and DETA with different frozen stages based on ResNet-50 backbone. The frozen stage "1" means only freezing the stem in the backbone. Frozen stage "2" means freezing the stem and the first residual stage and stage "0" means there is no frozen layer in the backbone.}
\vspace{-0mm}
{
\resizebox{1\textwidth}{!}{%
\begin{tabular}{c|c|c|c|ccccc}
\shline
\textbf{Model}  & \textbf{\#ep} & \textbf{\#frozen stages} & \textbf{AP} & \textbf{AP$_{50}$} &\textbf{ AP$_{75}$} & \textbf{AP$_{S}$} & \textbf{AP$_{M}$} & \textbf{AP$_{L}$} \\
\shline
DINO-4scale  & 12  & 0 & \textbf{49.7} & \textbf{67.0} & \textbf{54.4} & 31.4 & \textbf{52.9} & \textbf{63.6}  \\
DINO-4scale  & 12  & 1 & $49.4$ & 66.8 & 53.9 & 32.2 & 52.6 & 63.0  \\
DINO-4scale  & 12  & 2 & $49.4$ & 66.5 & 54.1 & \textbf{32.9} & 52.5 & 63.3 \\
\shline
DETA  & 12  & 0 & 49.8 & \textbf{67.5} & 54.7 & 31.9 & 54.2 & \textbf{65.2} \\
DETA  & 12  & 1 & \textbf{50.2} & 67.4 & \textbf{55.2} & \textbf{32.3} & \textbf{54.2} & 65.0  \\
DETA  & 12  & 2 & 49.9 & 67.3 & 54.7 & 32.1 & 54.2 & 64.8  \\
\shline
\end{tabular}}
\vspace{-0.2cm}
\label{tab:ablation_frozen_stage}
}
\end{table*}
  \begin{table*}[ht]
\centering\setlength{\tabcolsep}{5pt}
\renewcommand{\arraystretch}{1.3}
\footnotesize
\vspace{-0.2cm}
\caption{Ablation studies on hyper-parameters for DETR variants. For a fair comparison, we use ResNet-50 as the default backbone and \textbf{freeze the stem} in the backbone, which refers to "\#fronzen stages=1" in Table~\ref{tab:ablation_frozen_stage}.}
{
\resizebox{1.0\textwidth}{!}{%
\begin{tabular}{c|c|c|c|c|c|c}
\shline
\multirow{2}{*}{\textbf{Model}}  & \multirow{2}{*}{\textbf{\#ep}}  & \multicolumn{3}{c|}{\textbf{Learning rate}} & \multirow{2}{*}{\textbf{Class weight}} & \multirow{2}{*}{\textbf{AP}}  \\ \cline{3-5}
 & & \textbf{backbone} & \textbf{sampling-offsets \& reference-points} & \textbf{encoder-decoder} &  &   \\
\shline
Deformable-DETR-Two-Stage  & 50  & 2e-5  & 2e-5 & 2e-4 & 1.0 & 48.0 \\
Deformable-DETR-Two-Stage  & 50  & 2e-5  & 2e-5 & 2e-4 & 2.0 & 47.0 \\
Deformable-DETR-Two-Stage  & 50  & 1e-5  & 1e-4 & 1e-4 & 1.0 & \textbf{48.2} \\
Deformable-DETR-Two-Stage  & 50  & 1e-5  & 1e-4 & 1e-4 & 2.0 & 46.9 \\
\shline
$\mathcal{H}$-DETR  &  12  & 2e-5  & 2e-5 & 2e-4 & 1.0 & 48.9 \\
$\mathcal{H}$-DETR  &  12  & 2e-5  & 2e-5 & 2e-4 & 2.0 & \textbf{49.1} \\
$\mathcal{H}$-DETR  & 12 & 1e-5  & 1e-4 & 1e-4 & 1.0 & 48.5 \\
$\mathcal{H}$-DETR  &  12   & 1e-5  & 1e-4 & 1e-4 & 2.0 & 48.4 \\
\shline
DINO  &  12  & 2e-5  & 2e-5 & 2e-4 & 1.0 & 49.4 \\
DINO  &  12  & 2e-5  & 2e-5 & 2e-4 & 2.0 & \textbf{49.7} \\
DINO  &  12  & 1e-5  & 1e-4 & 1e-4 & 1.0 & 49.0 \\
DINO  &  12  & 1e-5  & 1e-4 & 1e-4 & 2.0 & 49.0 \\
\shline
\end{tabular}}
\label{tab:ablation_learning_rate}
}
\end{table*}
  \begin{table*}[ht]
\centering\setlength{\tabcolsep}{10pt}
\renewcommand{\arraystretch}{1.3}
\footnotesize
\vspace{-0.2cm}
{
\resizebox{1\textwidth}{!}{%
\begin{tabular}{l|c|c|ccccc}
\shline
\textbf{Model}  & \textbf{\#ep}  & \textbf{AP} & \textbf{AP$_{50}$} & \textbf{AP$_{75}$} & \textbf{AP$_{S}$} & \textbf{AP$_{M}$} & \textbf{AP$_{L}$} \\
\shline
Deformable-DETR  & 50   & $44.5$ & 63.6 & 48.7 & 27.0 & 47.6 & 59.6  \\
 \rowcolor{Gray}
Deformable-DETR (\textbf{\textcolor{rose}{detrex}})  & 50 & \textbf{44.9} \textbf{\textcolor{red}{(+0.4)}} & 64.0 & 48.6 & 26.8 & 48.1 & 60.2  \\
\shline
Deformable-DETR-Two-Stage  & 50   & $47.1$ & 66.0 & 51.3 & 30.9 & 50.2 & 61.9  \\
 \rowcolor{Gray}
Deformable-DETR-Two-Stage (\textbf{\textcolor{rose}{detrex}})  & 50 & \textbf{48.2} \textbf{\textcolor{red}{(+1.1)}} & 67.0 & 52.2 & 30.7 & 51.4 & 63.0  \\
\shline
Conditional-DETR  & 50   & 40.9 & 61.7 & 43.3 & 20.5 & 44.2 & 59.5  \\
 \rowcolor{Gray}
Conditional-DETR (\textbf{\textcolor{rose}{detrex}})  & 50 & \textbf{41.6 \textcolor{red}{(+0.7)}} & 63.0 & 43.9 & 21.4 & 45.2 & 59.8  \\
\shline
DAB-DETR  & 50   & $42.2$ & 63.1 & 44.7 & 21.5 & 45.7 & 60.3  \\
 \rowcolor{Gray}
DAB-DETR (\textbf{\textcolor{rose}{detrex}})  & 50 & \textbf{43.3 \textcolor{red}{(+1.1)}} & 63.9 & 45.9 & 23.4 & 47.1 & 62.1  \\
\shline
DAB-Deformable-DETR  & 50   & $48.7$ & 67.2 & 53.0 & 31.4 & 51.6 & 63.9  \\
 \rowcolor{Gray}
DAB-Deformable-DETR (\textbf{\textcolor{rose}{detrex}})  & 50 & \textbf{49.0 \textcolor{red}{(+0.3)}} & 67.4 & 53.4 & 31.5 & 52.1 & 64.4  \\
 \rowcolor{Gray}
DAB-Deformable-DETR-Two-Stage (\textbf{\textcolor{rose}{detrex}})  & 50 & \textbf{49.7 \textcolor{red}{(+1.0)}} & 68.0 & 54.3 & 31.9 & 53.2 & 64.7  \\
\shline
DN-DETR  & 50   & $44.4$ & 64.6 & 47.4 & 23.4 & 48.3 & 63.5 \\
 \rowcolor{Gray}
DN-DETR (\textbf{\textcolor{rose}{detrex}})  & 50 & \textbf{44.7 \textcolor{red}{(+0.3)}} & 65.3 & 47.5 & 23.7 & 48.7 & 64.1  \\
\shline
DINO  & 12   & $49.0$ & 66.5 & 53.5 & 32.3 & 52.4 & 63.7  \\
 \rowcolor{Gray}
DINO (\textbf{\textcolor{rose}{detrex}})  & 12 & \textbf{49.7 \textcolor{red}{(+0.7)}} & 67.0 & 54.4 & 31.4 & 52.9 & 63.6  \\
\shline
$\mathcal{H}$-DETR  & 12   & $48.6$ & 66.8 & 52.9 & 30.7 & 51.8 & 63.5 \\
 \rowcolor{Gray}
$\mathcal{H}$-DETR (\textbf{\textcolor{rose}{detrex}})  & 12 & \textbf{49.1 \textcolor{red}{(+0.5)}} & 66.9 & 53.7 & 32.2 & 52.3 & 63.8  \\
\shline
DETA  & 12   & $50.0$ & 67.4 & 55.0 & 31.6 & 54.1 & 65.4  \\
 \rowcolor{Gray}
DETA (\textbf{\textcolor{rose}{detrex}})  & 12 & \textbf{50.2 \textcolor{red}{(+0.2)}} & 67.4 & 55.2 & 32.3 & 54.2 & 65.0  \\
\shline
\end{tabular}}
\caption{Comparison the performance of DETR variants between \textbf{\textcolor{rose}{detrex}} implementations and their original implementations. }
\label{tab:comparison_detrex_with_original_implementation}
}
\vspace{-0.4cm}
\end{table*}
\section{Optimizing Detection Transformers}

  In this section, we primarily focus on the performance of models under the framework of \textbf{\textcolor{rose}{detrex}}. This evaluation encompasses model-sensitive hyper-parameters as well as the impact of post-processing techniques such as Non-Maximum Suppression (NMS) on DETR-based algorithms. Through meticulous optimization of the hyper-parameters, we successfully achieve superior model performance compared to the original implementations of DETR-based algorithms supported by \textbf{\textcolor{rose}{detrex}}.
  \vspace{-3mm}
  \paragraph{Ablation on evaluation with NMS.} We present the model evaluation results with NMS post-process in Table~\ref{tab:ablation_nms}. We find that post-processing with Non-maximum Suppression (NMS) can consistently improve the performance of DETR-variants, especially in terms of $\rm{AP}_{50}$ and $\rm{AP}_{L}$. We set the default threshold in NMS to 0.8, showing consistent gains on all models.
  \vspace{-3mm}
  \paragraph{Ablation on the training hyper-parameters.} In this subsection, we conduct a set of experiments to demonstrate that the DETR-variants are extremely sensitive to certain hyper-parameters, and we need to control them carefully. The DETR models can achieve significantly better performance by appropriately setting those hyper-parameters. 

  Firstly, we present a comparison of the effectiveness of different frozen backbone stages on DINO and DETA models in Table~\ref{tab:ablation_frozen_stage}. Under the same environment, the DINO model training without a frozen backbone and the DETA model training with only one frozen stage achieved the highest performance. A suitable backbone frozen stage can lead to a performance improvement of 0.3 AP.

  Furthermore, we conduct a detailed ablation study on commonly used hyper-parameters based on the Deformable-DETR, H-DETR, and DINO. Specifically, we investigate the learning rate configurations for different components of the models as well as the weight assigned to the classification loss in the loss function in Table~\ref{tab:ablation_learning_rate}. We discover that the models are highly sensitive to these commonly used hyper-parameters. For instance, in the case of Deformable-DETR-Two-Stage, by appropriately configuring the learning rate and weight of the classification loss, We have achieved a notable improvement of 1.3 AP (from 46.9 AP to 48.2 AP) on the final performance.

  \vspace{-3mm}
  \paragraph{Strong Detection Transformer Baselines.} From Table~\ref{tab:comparison_detrex_with_original_implementation}, our \textbf{\textcolor{rose}{detrex}} implementation achieves better performance across a broad range of DETR-based algorithms~\cite{liu2022dabdetr,zhu2020deformable, meng2021conditional, zhang2022dino, li2022dn, ouyang2022nms, jia2022detrs}. Through optimization of model hyper-parameters and training hyper-parameters, the Deformable-DETR and Deformable-DETR-Two-Stage models implemented under \textbf{\textcolor{rose}{detrex}} have achieved a 0.4 and 1.1 AP improvement, respectively, compared to their original implementations. The DAB-DETR surpasses its original implementation by 1.1 AP. Moreover, we have expanded the Two-Stage version of DAB-Deformable-DETR, which compared to the original implementation of DAB-Deformable-DETR, has achieved an improvement of 1.0 AP. Additionally, we boost the DINO and H-DETR models by an enhancement of 0.7 and 0.5 AP, respectively, over their original implementations.
\section{Conclusion}

We have presented \textbf{\textcolor{rose}{detrex}}, a benchmarking platform specifically designed for DETR-based models, addressing the lack of a unified and comprehensive benchmark in the field. 
By offering a highly modular and lightweight framework, \textbf{\textcolor{rose}{detrex}} supports a wide range of tasks and enables effective evaluation and comparison of DETR-based algorithms. 
Through extensive experiments, \textbf{\textcolor{rose}{detrex}} demonstrates improved model performance and provides valuable insights into the optimization of hyperparameters and the impact of different components. 
We hope that \textbf{\textcolor{rose}{detrex}} will serve as a standardized and consistent platform for the research community, fostering a deeper understanding and driving advancements in DETR-based models.

\clearpage
\bibliography{main}

\begin{thebibliography}{10}

\bibitem{bochkovskiy2020yolov4}
Alexey {Bochkovskiy}, Chien-Yao {Wang}, and Hong-Yuan~Mark {Liao}.
\newblock {YOLOv4: Optimal Speed and Accuracy of Object Detection}.
\newblock {\em arXiv preprint arXiv:2004.10934}, 2020.

\bibitem{carion2020end}
Nicolas Carion, Francisco Massa, Gabriel Synnaeve, Nicolas Usunier, Alexander
  Kirillov, and Sergey Zagoruyko.
\newblock {End-to-End Object Detection with Transformers}.
\newblock In {\em ECCV}, pages 213--229. Springer, 2020.

\bibitem{chen2019mmdetection}
Kai Chen, Jiaqi Wang, Jiangmiao Pang, Yuhang Cao, Yu~Xiong, Xiaoxiao Li,
  Shuyang Sun, Wansen Feng, Ziwei Liu, Jiarui Xu, et~al.
\newblock {MMDetection: Open MMLab Detection Toolbox and Benchmark}.
\newblock {\em arXiv preprint arXiv:1906.07155}, 2019.

\bibitem{chen2022group}
Qiang Chen, Xiaokang Chen, Gang Zeng, and Jingdong Wang.
\newblock Group detr: Fast training convergence with decoupled one-to-many
  label assignment.
\newblock {\em arXiv preprint arXiv:2207.13085}, 2022.

\bibitem{chen2022group2}
Qiang Chen, Jian Wang, Chuchu Han, Shan Zhang, Zexian Li, Xiaokang Chen, Jiahui
  Chen, Xiaodi Wang, Shuming Han, Gang Zhang, et~al.
\newblock Group detr v2: Strong object detector with encoder-decoder
  pretraining.
\newblock {\em arXiv preprint arXiv:2211.03594}, 2022.

\bibitem{cheng2022masked}
Bowen Cheng, Ishan Misra, Alexander~G Schwing, Alexander Kirillov, and Rohit
  Girdhar.
\newblock {Masked-attention Mask Transformer for Universal Image Segmentation}.
\newblock In {\em CVPR}, pages 1290--1299, 2022.

\bibitem{cheng2021maskformer}
Bowen Cheng, Alexander~G. Schwing, and Alexander Kirillov.
\newblock Per-pixel classification is not all you need for semantic
  segmentation.
\newblock {\em NeurIPS}, 2021.

\bibitem{cordts2016cityscapes}
Marius Cordts, Mohamed Omran, Sebastian Ramos, Timo Rehfeld, Markus Enzweiler,
  Rodrigo Benenson, Uwe Franke, Stefan Roth, and Bernt Schiele.
\newblock {The Cityscapes Dataset for Semantic Urban Scene Understanding}.
\newblock In {\em CVPR}, pages 3213--3223, 2016.

\bibitem{gao2021fast}
Peng Gao, Minghang Zheng, Xiaogang Wang, Jifeng Dai, and Hongsheng Li.
\newblock Fast convergence of detr with spatially modulated co-attention.
\newblock {\em arXiv preprint arXiv:2101.07448}, 2021.

\bibitem{ge2021yolox}
Zheng {Ge}, Songtao {Liu}, Feng {Wang}, Zeming {Li}, and Jian {Sun}.
\newblock {YOLOX: Exceeding YOLO Series in 2021}.
\newblock {\em arXiv preprint arXiv:2107.08430}, 2021.

\bibitem{he2015deep}
Kaiming {He}, Xiangyu {Zhang}, Shaoqing {Ren}, and Jian {Sun}.
\newblock {Deep Residual Learning for Image Recognition}.
\newblock In {\em 2016 IEEE Conference on Computer Vision and Pattern
  Recognition (CVPR)}, pages 770--778, 2016.

\bibitem{hu2021istr}
Jie Hu, Liujuan Cao, Yao Lu, ShengChuan Zhang, Yan Wang, Ke~Li, Feiyue Huang,
  Ling Shao, and Rongrong Ji.
\newblock {ISTR: End-to-End Instance Segmentation with Transformers}, 2021.

\bibitem{hu2023segment}
Jie Hu, Linyan Huang, Tianhe Ren, Shengchuan Zhang, Rongrong Ji, and Liujuan
  Cao.
\newblock {You Only Segment Once: Towards Real-Time Panoptic Segmentation},
  2023.

\bibitem{jia2022detrs}
Ding Jia, Yuhui Yuan, Haodi He, Xiaopei Wu, Haojun Yu, Weihong Lin, Lei Sun,
  Chao Zhang, and Han Hu.
\newblock {DETRs with Hybrid Matching}.
\newblock {\em arXiv preprint arXiv:2207.13080}, 2022.

\bibitem{kamath2021mdetr}
Aishwarya Kamath, Mannat Singh, Yann LeCun, Gabriel Synnaeve, Ishan Misra, and
  Nicolas Carion.
\newblock {MDETR - Modulated Detection for End-to-End Multi-Modal
  Understanding}.
\newblock In {\em ICCV}, pages 1780--1790, 2021.

\bibitem{law2018cornernet}
Hei Law and Jia Deng.
\newblock Cornernet: Detecting objects as paired keypoints.
\newblock In {\em ECCV}, pages 734--750, 2018.

\bibitem{li2022dn}
Feng Li, Hao Zhang, Shilong Liu, Jian Guo, Lionel~M Ni, and Lei Zhang.
\newblock {{DN}-{DETR}: Accelerate DETR Training by Introducing Query
  DeNoising}.
\newblock In {\em CVPR}, 2022.

\bibitem{li2022mask}
Feng Li, Hao Zhang, Shilong Liu, Lei Zhang, Lionel~M Ni, Heung-Yeung Shum,
  et~al.
\newblock {Mask DINO: Towards A Unified Transformer-based Framework for Object
  Detection and Segmentation}.
\newblock {\em arXiv preprint arXiv:2206.02777}, 2022.

\bibitem{li2022exploring}
Yanghao Li, Hanzi Mao, Ross Girshick, and Kaiming He.
\newblock {Exploring Plain Vision Transformer Backbones for Object Detection}.
\newblock In {\em ECCV}, pages 280--296. Springer, 2022.

\bibitem{li2022bevformer}
Zhiqi Li, Wenhai Wang, Hongyang Li, Enze Xie, Chonghao Sima, Tong Lu, Qiao Yu,
  and Jifeng Dai.
\newblock {BEVFormer: Learning Bird's-Eye-View Representation from Multi-Camera
  Images via Spatiotemporal Transformers}, 2022.

\bibitem{lin2014microsoft}
Tsung-Yi Lin, Michael Maire, Serge Belongie, James Hays, Pietro Perona, Deva
  Ramanan, Piotr Doll{\'a}r, and C~Lawrence Zitnick.
\newblock {Microsoft COCO: Common Objects in Context}.
\newblock In {\em ECCV}, pages 740--755. Springer, 2014.

\bibitem{liu2022dabdetr}
Shilong Liu, Feng Li, Hao Zhang, Xiao Yang, Xianbiao Qi, Hang Su, Jun Zhu, and
  Lei Zhang.
\newblock {{DAB}-{DETR}: Dynamic Anchor Boxes are Better Queries for {DETR}}.
\newblock In {\em ICLR}, 2022.

\bibitem{liu2022dq}
Shilong Liu, Yaoyuan Liang, Feng Li, Shijia Huang, Hao Zhang, Hang Su, Jun Zhu,
  and Lei Zhang.
\newblock {DQ-DETR: Dual Query Detection Transformer for Phrase Extraction and
  Grounding}.
\newblock {\em arXiv preprint arXiv:2211.15516}, 2022.

\bibitem{liu2023detection}
Shilong Liu, Tianhe Ren, Jiayu Chen, Zhaoyang Zeng, Hao Zhang, Feng Li,
  Hongyang Li, Jun Huang, Hang Su, Jun Zhu, et~al.
\newblock {Detection Transformer with Stable Matching}.
\newblock {\em arXiv preprint arXiv:2304.04742}, 2023.

\bibitem{liu2023grounding}
Shilong Liu, Zhaoyang Zeng, Tianhe Ren, Feng Li, Hao Zhang, Jie Yang, Chunyuan
  Li, Jianwei Yang, Hang Su, Jun Zhu, et~al.
\newblock {Grounding DINO: Marrying DINO with Grounded Pre-Training for
  Open-Set Object Detection}.
\newblock {\em arXiv preprint arXiv:2303.05499}, 2023.

\bibitem{liu2016ssd}
Wei Liu, Dragomir Anguelov, Dumitru Erhan, Christian Szegedy, Scott Reed,
  Cheng-Yang Fu, and Alexander~C Berg.
\newblock {SSD: Single Shot MultiBox Detector}.
\newblock In {\em ECCV}, pages 21--37. Springer, 2016.

\bibitem{liu2021Swin}
Ze~Liu, Yutong Lin, Yue Cao, Han Hu, Yixuan Wei, Zheng Zhang, Stephen Lin, and
  Baining Guo.
\newblock {Swin Transformer: Hierarchical Vision Transformer using Shifted
  Windows}.
\newblock {\em arXiv preprint arXiv:2103.14030}, 2021.

\bibitem{liu2022convnet}
Zhuang Liu, Hanzi Mao, Chao-Yuan Wu, Christoph Feichtenhofer, Trevor Darrell,
  and Saining Xie.
\newblock {A ConvNet for the 2020s}.
\newblock {\em CVPR}, 2022.

\bibitem{loshchilov2019decoupled}
Ilya Loshchilov and Frank Hutter.
\newblock {Decoupled Weight Decay Regularization}.
\newblock In {\em ICLR}, 2018.

\bibitem{lv2023detrs}
Wenyu Lv, Shangliang Xu, Yian Zhao, Guanzhong Wang, Jinman Wei, Cheng Cui,
  Yuning Du, Qingqing Dang, and Yi~Liu.
\newblock {DETRs Beat YOLOs on Real-time Object Detection}.
\newblock {\em arXiv preprint arXiv:2304.08069}, 2023.

\bibitem{meng2021conditional}
Depu Meng, Xiaokang Chen, Zejia Fan, Gang Zeng, Houqiang Li, Yuhui Yuan, Lei
  Sun, and Jingdong Wang.
\newblock {Conditional DETR for Fast Training Convergence}.
\newblock {\em arXiv preprint arXiv:2108.06152}, 2021.

\bibitem{neuhold2017mapillary}
Gerhard Neuhold, Tobias Ollmann, Samuel Rota~Bulo, and Peter Kontschieder.
\newblock {The Mapillary Vistas Dataset for Semantic Understanding of Street
  Scenes}.
\newblock In {\em ICCV}, pages 4990--4999, 2017.

\bibitem{nguyen2022boxer}
Duy-Kien Nguyen, Jihong Ju, Olaf Booij, Martin~R Oswald, and Cees~GM Snoek.
\newblock {BoxeR: Box-Attention for 2D and 3D Transformers}.
\newblock In {\em CVPR}, pages 4773--4782, 2022.

\bibitem{ouyang2022nms}
Jeffrey Ouyang-Zhang, Jang~Hyun Cho, Xingyi Zhou, and Philipp
  Kr{\"a}henb{\"u}hl.
\newblock {NMS Strikes Back}.
\newblock {\em arXiv preprint arXiv:2212.06137}, 2022.

\bibitem{redmon2016you}
Joseph Redmon, Santosh Divvala, Ross Girshick, and Ali Farhadi.
\newblock {You Only Look Once: Unified, Real-Time Object Detection}.
\newblock In {\em CVPR}, pages 779--788, 2016.

\bibitem{redmon2017yolo9000}
Joseph Redmon and Ali Farhadi.
\newblock {YOLO9000: Better, Faster, Stronger}.
\newblock In {\em CVPR}, pages 7263--7271, 2017.

\bibitem{ren2015faster}
Shaoqing {Ren}, Kaiming {He}, Ross {Girshick}, and Jian {Sun}.
\newblock {Faster R-CNN: Towards Real-Time Object Detection with Region
  Proposal Networks}.
\newblock {\em IEEE Transactions on Pattern Analysis and Machine Intelligence},
  39(6):1137--1149, 2017.

\bibitem{ren2023strong}
Tianhe Ren, Jianwei Yang, Shilong Liu, Ailing Zeng, Feng Li, Hao Zhang,
  Hongyang Li, Zhaoyang Zeng, and Lei Zhang.
\newblock {A Strong and Reproducible Object Detector with Only Public
  Datasets}, 2023.

\bibitem{shao2019objects365}
Shuai Shao, Zeming Li, Tianyuan Zhang, Chao Peng, Gang Yu, Xiangyu Zhang, Jing
  Li, and Jian Sun.
\newblock {Objects365: A Large-scale, High-quality Dataset for Object
  Detection}.
\newblock In {\em ICCV}, pages 8430--8439, 2019.

\bibitem{Shi_2022_CVPR}
Dahu Shi, Xing Wei, Liangqi Li, Ye~Ren, and Wenming Tan.
\newblock {End-to-End Multi-Person Pose Estimation With Transformers}.
\newblock In {\em CVPR}, pages 11069--11078, June 2022.

\bibitem{tian2019fcos}
Zhi {Tian}, Chunhua {Shen}, Hao {Chen}, and Tong {He}.
\newblock {FCOS: Fully Convolutional One-Stage Object Detection}.
\newblock In {\em 2019 IEEE/CVF International Conference on Computer Vision
  (ICCV)}, pages 9627--9636, 2019.

\bibitem{wang2022internimage}
Wenhai Wang, Jifeng Dai, Zhe Chen, Zhenhang Huang, Zhiqi Li, Xizhou Zhu,
  Xiaowei Hu, Tong Lu, Lewei Lu, Hongsheng Li, et~al.
\newblock {InternImage: Exploring Large-Scale Vision Foundation Models with
  Deformable Convolutions}.
\newblock {\em arXiv preprint arXiv:2211.05778}, 2022.

\bibitem{wang2021anchor}
Yingming {Wang}, Xiangyu {Zhang}, Tong {Yang}, and Jian {Sun}.
\newblock {Anchor DETR: Query Design for Transformer-Based Detector}.
\newblock {\em arXiv preprint arXiv:2109.07107}, 2021.

\bibitem{wu2019detectron2}
Yuxin Wu, Alexander Kirillov, Francisco Massa, Wan-Yen Lo, and Ross Girshick.
\newblock Detectron2.
\newblock \url{https://github.com/facebookresearch/detectron2}, 2019.

\bibitem{xie2021segformer}
Enze Xie, Wenhai Wang, Zhiding Yu, Anima Anandkumar, Jose~M Alvarez, and Ping
  Luo.
\newblock {SegFormer: Simple and Efficient Design for Semantic Segmentation
  with Transformers}.
\newblock {\em NeurIPS}, 34:12077--12090, 2021.

\bibitem{yang2022focal}
Jianwei Yang, Chunyuan Li, Xiyang Dai, and Jianfeng Gao.
\newblock {Focal Modulation Network}.
\newblock {\em NeurIPS}, 35:4203--4217, 2022.

\bibitem{yang2023boosting}
Jie Yang, Bingliang Li, Fengyu Yang, Ailing Zeng, Lei Zhang, and Ruimao Zhang.
\newblock {Boosting Human-Object Interaction Detection with Text-to-Image
  Diffusion Model}, 2023.

\bibitem{yang2023explicit}
Jie Yang, Ailing Zeng, Shilong Liu, Feng Li, Ruimao Zhang, and Lei Zhang.
\newblock {Explicit Box Detection Unifies End-to-End Multi-Person Pose
  Estimation}.
\newblock {\em arXiv preprint arXiv:2302.01593}, 2023.

\bibitem{zhang2022dino}
Hao Zhang, Feng Li, Shilong Liu, Lei Zhang, Hang Su, Jun Zhu, Lionel~M Ni, and
  Heung-Yeung Shum.
\newblock {DINO: DETR with Improved DeNoising Anchor Boxes for End-to-End
  Object Detection}.
\newblock {\em arXiv preprint arXiv:2203.03605}, 2022.

\bibitem{zhang2023mp}
Hao Zhang, Feng Li, Huaizhe Xu, Shijia Huang, Shilong Liu, Lionel~M Ni, and Lei
  Zhang.
\newblock {MP-Former: Mask-Piloted Transformer for Image Segmentation}.
\newblock In {\em CVPR}, pages 18074--18083, 2023.

\bibitem{zhang2023bev}
Hao Zhang, Hongyang Li, Xingyu Liao, Feng Li, Shilong Liu, Lionel~M Ni, and Lei
  Zhang.
\newblock {DA-BEV: Depth Aware BEV Transformer for 3D Object Detection}.
\newblock {\em arXiv preprint arXiv:2302.13002}, 2023.

\bibitem{zhang2023dense}
Shilong Zhang, Xinjiang Wang, Jiaqi Wang, Jiangmiao Pang, Chengqi Lyu, Wenwei
  Zhang, Ping Luo, and Kai Chen.
\newblock {Dense Distinct Query for End-to-End Object Detection}.
\newblock In {\em CVPR}, pages 7329--7338, 2023.

\bibitem{zhou2019semantic}
Bolei Zhou, Hang Zhao, Xavier Puig, Tete Xiao, Sanja Fidler, Adela Barriuso,
  and Antonio Torralba.
\newblock {Semantic Understanding of Scenes through the ADE20K Dataset}.
\newblock {\em IJCV}, 127:302--321, 2019.

\bibitem{zhu2020deformable}
Xizhou {Zhu}, Weijie {Su}, Lewei {Lu}, Bin {Li}, Xiaogang {Wang}, and Jifeng
  {Dai}.
\newblock {Deformable DETR: Deformable Transformers for End-to-End Object
  Detection}.
\newblock 2021.

\end{thebibliography}
\bibliographystyle{plain}





\end{document}